\documentclass[sigconf]{acmart}

\copyrightyear{2026}
\acmYear{2026}
\setcopyright{cc}

\setcctype{by}

\acmConference[MM '26]
{Proceedings of the 34th ACM International Conference on Multimedia}
{November 10--14, 2026}
{Rio de Janeiro, Brazil.}

\acmBooktitle{Proceedings of the 34th ACM International Conference on Multimedia
(MM '26), November 10--14, 2026, Rio de Janeiro, Brazil}

\acmISBN{979-8-4007-2213-4/2026/11}
\acmDOI{10.1145/3767308.3834921}

\AtBeginDocument{%
  }

\usepackage{multirow}
\usepackage{makecell}

\begin{document}

\title{Phase-Aligned Finite-Fourier Periodic Deformation for 4D Medical Image Interpolation}

\author{Haojin Li}
\authornote{Haojin Li and Hengzhuo Wang contributed equally to this work.}
\affiliation{%
  \department{Department of Computer Science and Engineering}
  \institution{Southern University of Science and Technology}
  \city{Shenzhen}
  \country{China}
}
\email{12232410@mail.sustech.edu.cn}

\author{Hengzhuo Wang}
\authornotemark[1]
\affiliation{%
  \department{Faculty of Biomedical Engineering}
  \institution{Shenzhen University of Advanced Technology}
  \city{Shenzhen}
  \country{China}
}
\affiliation{%
  \department{School of Biomedical Engineering, Shenzhen University Medical School}
  \institution{Shenzhen University}
  \city{Shenzhen}
  \country{China}
}
\email{suat25060145@stu.suat-sz.edu.cn}

\author{Zhiheng Ma}
\affiliation{%
  \department{Faculty of Computility Microelectronics}
  \institution{Shenzhen University of Advanced Technology}
  \city{Shenzhen}
  \country{China}
}
\email{mazhiheng@suat-sz.edu.cn}

\author{Mingyang Ou}
\affiliation{%
  \department{School of Computer Science and Engineering}
  \institution{Beihang University}
  \city{Beijing}
  \country{China}
}
\email{ComgLqWork@outlook.com}

\author{Heng Li}
\authornote{Heng Li and Jiang Liu are the corresponding authors.}
\affiliation{%
  \department{Faculty of Biomedical Engineering}
  \institution{Shenzhen University of Advanced Technology}
  \city{Shenzhen}
  \country{China}
}
\email{liheng@suat-sz.edu.cn}

\author{Jiang Liu}
\authornotemark[2]
\affiliation{%
  \department{Department of Computer Science and Engineering}
  \institution{Southern University of Science and Technology}
  \city{Shenzhen}
  \country{China}
}
\email{liuj@sustech.edu.cn}

\renewcommand{\shortauthors}{Haojin Li et al.}


\begin{abstract}
4D medical image interpolation aims to recover missing volumes from sparsely observed time points and is important for dynamic anatomical analysis in applications such as cardiac MRI and thoracic CT, where motion is often repetitive or near-periodic over clinically relevant intervals. A key challenge is that this structure is not always encoded directly in deformation representations for interpolation. In addition, physiological motion is often non-uniform, so equal temporal intervals do not necessarily correspond to equal amounts of anatomical change. To address these issues, we formulate interpolation as learning a continuous deformation process with a phase-structured prior. Given two endpoint volumes, we parameterize a phase-conditioned velocity field with a finite Fourier basis, which embeds near-periodic motion patterns directly into the deformation space and supports continuous querying at arbitrary target times. We further introduce a phase-aligned temporal reparameterization that maps normalized within-interval time to a latent motion phase according to deformation variation intensity, thereby better modeling non-uniform motion progression. Intermediate volumes are then synthesized by continuously warping both endpoints, followed by bidirectional fusion and lightweight residual refinement. Experiments on ACDC and 4D-Lung show that the proposed method achieves state-of-the-art performance over existing baselines while producing anatomically plausible and coherent intermediate volumes from sparse observations.

\end{abstract}



\maketitle

\section{Introduction}

4D medical imaging is widely used in functional assessment, treatment monitoring, and dynamic anatomical analysis~\cite{yan2025online}. Representative applications include cardiac motion analysis in cine MRI and respiratory motion analysis in 4D thoracic CT, where clinically meaningful information is determined not only by spatial appearance but also by anatomical deformation over time~\cite{guo2020spatiotemporal,li2024cpt,wang2025canfields}. In practice, however, dynamic acquisitions are often limited by low temporal resolution, sparse sampling, or irregular acquisition intervals, leaving a substantial portion of the motion process unobserved~\cite{perrin2025super,li2025unsupervised}. As a result, 4D medical image interpolation is both practically important and technically challenging. The task must recover missing volumes from only a small number of observed time points while preserving the continuity and anatomical consistency of the underlying motion, which in many dynamic imaging settings also exhibits partially repeatable or near-periodic patterns~\cite{chan2021full,dong2025multicycle}.

Existing 4D interpolation methods mainly fall into two paradigms. Deformation-based methods estimate motion, flow, or deformation between observed frames and reconstruct intermediate states through inferred correspondences~\cite{guo2020spatiotemporal,wei2023mpvf,kim2024data}, while direct synthesis methods generate missing frames more directly from image features together with temporal conditions~\cite{kim2022diffusion,you2025fb,zhang2025temporal,zhou2025diffusion}. Some methods further incorporate temporal regularity or periodic cues, for example through temporal conditioning or auxiliary guidance, to improve interpolation under structured motion~\cite{guo2020spatiotemporal,zhang2025temporal,li2024cpt,wang2025canfields}. However, such temporal structure is still usually introduced to support reconstruction rather than encoded in the deformation representation itself~\cite{li2024fld}. For 4D medical imaging, where anatomical continuity and dynamic consistency are central, it is therefore more desirable to encode such clinically meaningful temporal structure directly within the underlying deformation process.

This leads to two key challenges. (1) How can we build a continuous deformation representation that explicitly captures structured motion? In many clinically important dynamic imaging applications, the underlying motion may exhibit partially repeatable or near-periodic behavior~\cite{guo2020spatiotemporal,yang2025latent,lu2025optical}. If such structure is used only as temporal conditioning for reconstructing a queried target volume, it does not directly constrain the deformation process, making it difficult to obtain a representation that remains continuous and can be queried reliably at arbitrary target times. (2) How can we account for uneven motion progression within a physiological cycle? Even when the overall motion is roughly repetitive, anatomical change does not evolve at a uniform pace within a cycle, so equal observation-time intervals may still correspond to different amounts of deformation. This makes linear time a suboptimal coordinate for continuous interpolation~\cite{yang2025latent,lu2025optical,hadji2021representation,cho2023neural}.

To address these challenges, we reformulate 4D medical image interpolation as learning a structured deformation process that can be queried continuously from sparse endpoint observations, rather than predicting intermediate states independently at queried time points. Our framework is built on two key ideas. First, we represent the motion process as a phase-conditioned continuous velocity field parameterized by a finite Fourier basis, so that repetitive or near-periodic structure is encoded directly in the deformation representation. Second, we introduce a phase-aligned temporal reparameterization that maps queried time within the endpoint interval to a phase coordinate according to deformation variation intensity, thereby better reflecting physiological dynamics that do not always evolve in proportion to physical time. Based on this phase-aware deformation process, an intermediate volume at any target time is synthesized by bidirectional endpoint warping, followed by fusion and lightweight residual refinement. This motion-centric formulation enables continuous interpolation that is better aligned with the underlying anatomical dynamics.

In summary, this work makes the following contributions:
\begin{itemize}
    \item We present a framework for 4D medical image interpolation based on phase-structured continuous motion modeling, enabling anatomically coherent intermediate synthesis from sparse endpoint observations at arbitrary query times.
    \item We introduce a finite-Fourier parameterization of a phase-conditioned continuous velocity field to encode repetitive or near-periodic motion structure directly in the deformation representation.
    \item We introduce a time-to-phase reparameterization mechanism that aligns queried time within the endpoint interval with motion phase progression, thereby better capturing non-uniform motion evolution.
    \item We validate the proposed method on the ACDC and 4D-Lung benchmarks, where it achieves state-of-the-art interpolation performance, and ablation studies further verify the distinct contributions of phase-structured deformation modeling and temporal reparameterization.
\end{itemize}

\section{Related Works}

\subsection{Video Frame Interpolation}

Video frame interpolation has been extensively studied in natural scenes and mainly follows two paradigms. Deformation-based methods estimate optical flow or related correspondences and synthesize intermediate frames through warping and refinement, with designs addressing large-motion matching, intermediate motion estimation, and trajectory-aware synthesis \cite{li2023amt,kong2022ifrnet,yu2023range,hu2024iq}. Direct synthesis methods formulate interpolation as conditional generation to improve flexibility and visual realism in ambiguous regions \cite{danier2024ldmvfi}. Recent work further introduces temporal regularity, arbitrary-time interpolation, and continuous video modeling for non-uniform motion and flexible querying \cite{he2022timereplayer,shrivastava2024video,zhang2025continuous}. These developments benefit motion estimation and continuous-time prediction, but primarily target visually plausible synthesis in natural scenes. By contrast, 4D medical image interpolation must also preserve anatomy-constrained deformation consistency under structured physiological motion.

\subsection{4D Medical Image Interpolation}

Research on 4D medical image interpolation has increasingly focused on deformable anatomical modeling. The main route is deformation-based interpolation, which estimates volumetric motion or voxel flow between observed scans and reconstructs intermediate volumes through warping \cite{guo2020spatiotemporal,wei2023mpvf,kim2024data}. This direction is closely related to learning-based deformable registration, where stronger backbones improve dense correspondence estimation and continuous or diffeomorphic parameterizations encourage smoother, anatomically plausible motion fields \cite{balakrishnan2019voxelmorph,chen2022transmorph,joshi2023r2net,vercauteren2009diffeomorphic}. Some studies introduce anatomical priors by representing motion on anatomical objects or surfaces \cite{meng2023deepmesh}. Others explore compact deformation parameterizations, such as band-limited Fourier representations, to regularize spatial motion structure \cite{jia2023fourier}. Together, these studies suggest that medical interpolation depends not only on reconstructing appearance but also on learning anatomically meaningful deformation.

\begin{figure*}[tp]
\includegraphics[width=0.9\textwidth]{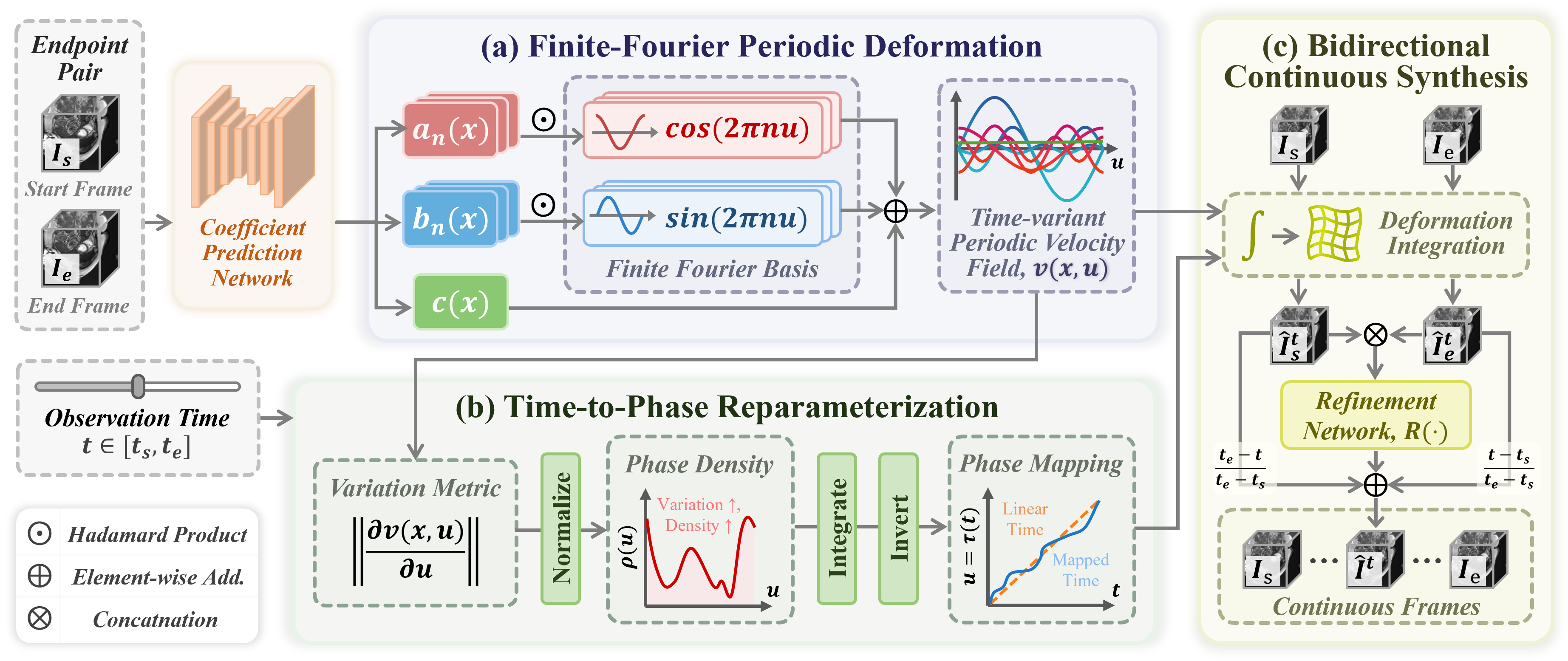}
\caption{Overview of the proposed framework. (a) The endpoint pair $(I_{\mathrm{s}}, I_{\mathrm{e}})$ is used to predict Fourier coefficient fields that parameterize a phase-conditioned periodic velocity field $v(x,u)$. (b) The queried time $t$ is reparameterized to phase $u=\tau(t)$ according to deformation variation. (c) The volume at time $t$ is synthesized by bidirectional deformation integration, followed by fusion and refinement.}
\label{fig:framework}
\Description{The figure illustrates the overall framework for continuous 4D medical image interpolation, consisting of three main components: (a) finite-Fourier periodic deformation, (b) time-to-phase reparameterization, and (c) bidirectional continuous synthesis.

On the left, the inputs include an endpoint pair of volumes, denoted as the start frame $I_s$ and end frame $I_e$, together with an observation time $t \in [t_s, t_e]$. These inputs are fed into a coefficient prediction network, which outputs spatially varying Fourier coefficients $a_n(x)$, $b_n(x)$, and a constant term $c(x)$.

In module (a), these coefficients are combined with cosine and sine basis functions, $\cos(2\pi n u)$ and $\sin(2\pi n u)$, where $u$ denotes a normalized motion phase. Through element-wise multiplication and summation, a time-varying periodic velocity field $v(x,u)$ is constructed using a finite Fourier basis. This field represents continuous deformation dynamics over the phase domain.

In module (b), a time-to-phase reparameterization is performed to handle non-uniform motion progression. First, a variation metric $\|\partial v(x,u)/\partial u\|$ is computed to measure deformation intensity over phase. This is normalized to obtain a phase density function $\rho(u)$, where regions with larger deformation correspond to higher density. The density is then integrated and inverted to produce a mapping from observation time $t$ to phase $u=\tau(t)$, aligning physical time with deformation progression.

In module (c), bidirectional continuous synthesis is performed. The phase-aligned velocity field is integrated over time to obtain deformation fields. Both endpoint volumes $I_s$ and $I_e$ are warped toward the target time using deformation integration. The resulting intermediate estimates are combined and refined through a lightweight refinement network $R(\cdot)$. The final output is a sequence of continuous intermediate frames between $I_s$ and $I_e$.

Symbols in the figure indicate operations: circles denote Hadamard (element-wise) products, circled plus signs denote element-wise addition, and crossed circles denote concatenation.}
\end{figure*}

Beyond this deformation-centered route, 4D medical interpolation methods have also introduced temporal regularity and synthesis-oriented designs for reconstruction under complex dynamics. Earlier spatiotemporal volumetric models already considered temporal context and periodic motion cues \cite{guo2020spatiotemporal}, while more recent approaches combine deformation modeling with diffusion-based generation, multi-scale fusion, temporal modulation, or Fourier-guided synthesis to handle nonlinear dynamics and large anatomical variation \cite{kim2022diffusion,you2025fb,zhang2025temporal,zhou2025diffusion}. Continuous motion formulations further extend the problem from prediction at isolated timestamps to recovery of a continuously queryable spatiotemporal process \cite{li2024cpt,wang2025canfields}. Nevertheless, in most existing methods, temporal regularity or periodic information is still used mainly as conditioning or auxiliary guidance. It is rarely encoded more directly in the deformation representation itself, leaving open the problem of learning a continuous anatomical deformation process whose representation can better capture near-periodic structure.

\subsection{Continuous Temporal Modeling}

Continuous temporal modeling has emerged in medical imaging and adjacent sequence modeling as an alternative to discrete frame indexing. Continuous spatiotemporal motion models, neural implicit motion fields, and manifold-based dynamic reconstruction methods enable arbitrary-time querying by treating the underlying dynamics as a continuous process \cite{li2024cpt,shen2024continuous,yoo2021time,hamilton2025multifrequency,banus2025nimosef}. Related work also shows that temporal coordinates may require explicit alignment. Phase detection methods model latent cardiac motion progression \cite{yang2025latent,lu2025optical}, while broader studies consider differentiable time warping, temporal synchronization, and global alignment across sequences with different traversal speeds \cite{cuturi2017soft,hadji2021representation,cho2023neural,betancur2015synchronization}. However, these methods mainly focus on query timing, phase detection, or sequence alignment. They do not explicitly reparameterize interpolation time by motion phase or unify phase-aware alignment with a near-periodic continuous deformation representation. Our method addresses this gap by mapping queried time within the observed interval to a phase coordinate that accounts for non-uniform motion progression.

\section{Methodology}

\subsection{Task Formulation and Framework Overview}

Let $\Omega \subset \mathbb{R}^3$ denote the spatial domain and $\{I_t:\Omega\rightarrow\mathbb{R}\}$ a dynamic volumetric sequence. In the endpoint-conditioned setting, each sample provides two observed volumes $I_s := I_{t_s}$ and $I_e := I_{t_e}$ at times $t_s < t_e$. Given $(I_s, I_e)$ and any queried time $t \in [t_s, t_e]$, the goal is to recover the target volume continuously between the two observations. Formally, the task is to learn a continuous interpolation function:
\begin{equation}
\hat{I}^{t} = \mathcal{F}_{\theta}(I_{\mathrm{s}}, I_{\mathrm{e}}, t), \qquad t \in [t_{\mathrm{s}}, t_{\mathrm{e}}],
\label{eq:task_definition_continuous}
\end{equation}
where $\hat{I}^{t}$ denotes the predicted volume at time $t$ and $\theta$ denotes the model parameters. During training, a sample may additionally provide intermediate observations at timestamps $\mathcal{T} = \{t_k\}_{k=1}^{K-1} \subset (t_{\mathrm{s}}, t_{\mathrm{e}})$, such that the sample can be written as $(I_{\mathrm{s}}, I_{\mathrm{e}}, \{(t_k, I_{t_k})\}_{k=1}^{K-1})$. These intermediate volumes are used only as supervision and are not fed into the network.

A generic endpoint-conditioned learning objective is written as:
\begin{equation}
\theta^{\ast}
=
\arg\min_{\theta}
\;
\mathbb{E}
\left[
\frac{1}{|\mathcal{T}|}
\sum_{t \in \mathcal{T}}
\mathcal{L}_{\mathrm{rec}}(\hat{I}^{t}, I_t)
+
\lambda \, \mathcal{R}(m^{t})
\right],
\label{eq:task_definition_training}
\end{equation}
where $m^{t}$ denotes the motion representation at time $t$, such as a deformation or velocity field, and $\mathcal{R}(\cdot)$ regularizes its smoothness or physical plausibility.

Rather than instantiating $\mathcal{F}_{\theta}$ as a direct image-space regressor, our method models interpolation through a continuous deformation process. The queried time determines the target position within the endpoint interval, while a latent phase coordinate $u \in [0,1]$ serves as a representation coordinate for motion progression on a normalized phase axis. This separation decouples temporal localization from motion representation and forms the basis of the proposed method.

\subsection{Finite-Fourier Periodic Deformation}

We model endpoint-conditioned interpolation as a continuously queryable deformation process on a latent phase domain, rather than as a collection of independent predictions at isolated query times. This process is parameterized by a finite Fourier basis, which allows repetitive or near-periodic motion structure to be encoded directly in the representation. Specifically, let $v(x,u) \in \mathbb{R}^{3}$ denote a phase-conditioned velocity field, where $x \in \Omega$ and $u \in [0,1]$ indexes motion progression on a normalized phase interval. Under this formulation, intermediate synthesis is derived from a unified anatomical evolution process rather than from disconnected image-space predictions.

To encode phase-wise regularity explicitly, we represent $v(x,u)$ by a finite Fourier expansion on the phase axis:
\begin{equation}
v(x,u)
=
c(x)
+
\sum_{n=1}^{N}
\left[
a_n(x)\cos(2\pi n u)
+
b_n(x)\sin(2\pi n u)
\right],
\label{eq:fourier_periodic_velocity}
\end{equation}
where $N$ denotes the truncation order, $c(x) \in \mathbb{R}^{3}$ is the zeroth-order component, and $a_n(x), b_n(x) \in \mathbb{R}^{3}$ are the cosine and sine coefficient fields at order $n$. This gives $v(x,u+1)=v(x,u)$ by construction, imposing periodicity at the representation level. Lower-order components capture coarse phase-scale motion, while higher-order terms model finer phase-dependent variation.

This formulation differs from both discrete interpolation and generic continuous regression. In our case, phase dependence is restricted to a finite Fourier subspace, making the periodic prior intrinsic to the admissible deformation family. The model therefore learns coefficients of a structured motion process rather than a collection of temporally indexed deformation snapshots.

In implementation, the network predicts the coefficient volumes $\{c,a_n,b_n\}_{n=1}^{N}$ from the endpoint pair $(I_{\mathrm{s}}, I_{\mathrm{e}})$ only once, after which the velocity field can be evaluated continuously at any phase $u$. To stabilize this truncated spectral representation, we impose frequency-aware regularization on the coefficient fields:
\begin{equation}
\mathcal{L}_{\mathrm{reg}}
=
\sum_{n=1}^{N}
n^{\gamma}
\left(
\|a_n\|_{L^{2}(\Omega)}^{2}
+
\|b_n\|_{L^{2}(\Omega)}^{2}
\right),
\label{eq:spectral_regularization}
\end{equation}
where $\gamma > 0$ controls the penalty strength on higher-order terms. We set $\gamma = 1.5$ in practice. This regularizer controls the spectral complexity of the phase-varying deformation process and improves the stability of continuous interpolation under finite-order truncation~\cite{ji2025pseudo}.

An additional advantage of this representation is that phase variation admits a direct spectral interpretation, which motivates the temporal alignment developed next.

\begin{proposition}[Spectral identity for phase variation]
\label{prop:spectral_identity_phase_variation}
For the finite-Fourier representation in Eq.~\eqref{eq:fourier_periodic_velocity}, the cycle-averaged squared phase variation satisfies:
\begin{equation}
\int_{0}^{1}\int_{\Omega}
\left\|
\partial_{u} v(x,u)
\right\|_{2}^{2}
\, dx \, du
=
2\pi^{2}
\sum_{n=1}^{N}
n^{2}
\left(
\|a_n\|_{L^{2}(\Omega)}^{2}
+
\|b_n\|_{L^{2}(\Omega)}^{2}
\right).
\label{eq:spectral_identity_phase_variation}
\end{equation}
\end{proposition}

This identity highlights an important advantage of the finite-Fourier deformation representation: phase-wise deformation variation is characterized explicitly by a frequency-squared weighted spectral energy. The contribution of each Fourier component to motion variation is therefore transparent, with higher-order components governing more rapidly changing phase-dependent dynamics. As a result, the learned spectrum admits a clear interpretation, and the temporal complexity of the deformation process can be analyzed directly from the coefficients rather than only being reflected implicitly in network activations. Proofs of this proposition, together with the derivations of subsequent related formulas, are provided in the supplementary material.

\subsection{Time-to-Phase Reparameterization}

Physiological motion generally does not evolve uniformly over observation time~\cite{chittajallu2018image}. Although the deformation process is represented continuously on the latent phase axis, equal increments in queried time do not necessarily correspond to equal progression along phase. Directly equating queried time with phase may therefore allocate temporal resolution poorly along the motion progression. We thus introduce a deterministic temporal reparameterization that maps queried time to a motion-aware phase coordinate while preserving the periodic deformation model.

We quantify phase-wise deformation variation by:
\begin{equation}
s(u)
=
\frac{1}{|\Omega|}
\int_{\Omega}
\left\|
\partial_{u} v(x,u)
\right\|_{2}^{2}
\, dx,
\qquad u \in [0,1].
\label{eq:phase_variation_metric}
\end{equation}

Based on this variation profile, we define the normalized phase density as:
\begin{equation}
\rho(u)
=
\frac{\varepsilon + s(u)}
{\int_{0}^{1} \left( \varepsilon + s(\xi) \right) d\xi},
\qquad \varepsilon = 10^{-6},
\label{eq:phase_density}
\end{equation}
where $\varepsilon$ is introduced to ensure that $\rho(u)$ remains strictly positive. The associated cumulative phase measure is then:
\begin{equation}
\Psi(u)
=
\int_{0}^{u} \rho(\xi)\, d\xi.
\label{eq:cumulative_phase_measure}
\end{equation}

The aligned phase coordinate is obtained by:
\begin{equation}
\tau(t)
=
\Psi^{-1}(t),
\qquad t \in [0,1].
\label{eq:phase_aligned_map}
\end{equation}

Since $\rho(u)$ is normalized on $[0,1]$, the cumulative map $\Psi$ satisfies $\Psi(0)=0$ and $\Psi(1)=1$. Therefore, its inverse $\tau=\Psi^{-1}$ is endpoint-aligned by construction, i.e., $\tau(0)=0$ and $\tau(1)=1$.

\begin{lemma}[Well-posedness of the phase-aligned map]
\label{lem:phase_aligned_map_wellposed}
Assume that $s$ is continuous on $[0,1]$. Then $\rho$ is continuous and strictly positive on $[0,1]$, $\Psi$ is a strictly increasing bijection from $[0,1]$ onto $[0,1]$, and $\tau=\Psi^{-1}$ is a continuous strictly increasing bijection satisfying $\tau(0)=0$ and $\tau(1)=1$. If, in addition, there exist constants $0<m\leq M<\infty$ such that
\begin{equation}
m \leq \rho(u) \leq M,
\qquad \forall u \in [0,1],
\label{eq:phase_density_bounds}
\end{equation}
then $\tau$ is bi-Lipschitz on $[0,1]$.
\end{lemma}

The reparameterization is induced by the learned deformation process itself rather than by an additional temporal predictor. Phases with larger deformation variation receive greater density mass and are therefore assigned finer temporal resolution after inversion by $\Psi$, whereas slowly varying phases are traversed more coarsely.

Let $V(u)=v(\cdot,u)$ denote the phase-indexed deformation trajectory in $L^{2}(\Omega;\mathbb{R}^{3})$, and let $\widetilde{V}(t)=V(\tau(t))$ denote its reparameterized counterpart. The following result shows that temporal reparameterization changes traversal speed but not the underlying trajectory.

\begin{theorem}[Uniform traversal under the phase metric]
\label{thm:uniform_traversal_phase_metric}
Assume that $V$ is absolutely continuous on $[0,1]$, and let $\tau$ be defined by Eqs.~\eqref{eq:phase_density}--\eqref{eq:phase_aligned_map}. Then $\widetilde{V}$ and $V$ have the same image:
\begin{equation}
\left\{
\widetilde{V}(t) : t \in [0,1]
\right\}
=
\left\{
V(u) : u \in [0,1]
\right\}.
\label{eq:same_trajectory_under_reparameterization}
\end{equation}
Moreover, $\widetilde{V}$ traverses this trajectory at constant speed with respect to the phase density, in the sense that
\begin{equation}
\rho(\tau(t))\,\tau'(t)
=
1
\qquad \text{for almost every } t \in [0,1].
\label{eq:constant_speed_under_phase_density}
\end{equation}
\end{theorem}

This result shows that the proposed mapping preserves the continuous periodic deformation trajectory while redistributing queried times according to phase-wise deformation complexity. The aligned phase coordinate $\tau(t)$ is subsequently used to evaluate the target deformation state for continuous synthesis from the two endpoints.

\subsection{Bidirectional Continuous Synthesis and Training Objective}

With the phase-aligned mapping $\tau(t)$, the learned periodic deformation process can be queried continuously at arbitrary observation times. Accordingly, we synthesize the target volume by transporting both endpoints to the queried time rather than by direct intensity regression. Specifically, let $W(I_{s}; s \rightarrow t)$ denote the volume obtained by warping a source volume $I_{s}$ from time $s$ to time $t$ along the deformation flow induced by the phase-conditioned velocity field evaluated on $\tau(\cdot)$, where the transport is computed by numerically integrating the corresponding flow with a discretized Euler scheme~\cite{sun2024medical}. The two directional candidates therefore differ only in their source endpoint and transport direction:
\begin{equation}
\hat{I}^{t}_{\mathrm{s}} = W(I_{\mathrm{s}}; t_{\mathrm{s}} \rightarrow t),
\qquad
\hat{I}^{t}_{\mathrm{e}} = W(I_{\mathrm{e}}; t_{\mathrm{e}} \rightarrow t).
\end{equation}

To ensure that the deformation-driven synthesis remains anatomically meaningful before refinement, we first supervise the two warped candidates directly. The morphing loss is defined as:
\begin{equation}
\mathcal{L}_{\mathrm{morph}}
=
\frac{1}{|\mathcal{T}|}
\sum_{t \in \mathcal{T}}
\left[
\mathcal{L}_{\mathrm{sim}}(\hat{I}_{\mathrm{s}}^{\,t}, I_t)
+
\mathcal{L}_{\mathrm{sim}}(\hat{I}_{\mathrm{e}}^{\,t}, I_t)
\right],
\label{eq:morph_loss_34}
\end{equation}
where $\mathcal{T}$ denotes the set of supervised timestamps, and $\mathcal{L}_{\mathrm{sim}}(\cdot,\cdot)=\mathcal{L}_{\mathrm{ncc}}(\cdot,\cdot)+\lambda_{\mathrm{charb}}\mathcal{L}_{\mathrm{charb}}(\cdot,\cdot)$ combines a local normalized cross-correlation loss with a Charbonnier term weighted by a small coefficient $\lambda_{\mathrm{charb}}$.

In addition, the learned periodic transport should remain consistent over a full normalized phase traversal. We therefore construct a forward cycle by transporting $I_{\mathrm{s}}$ through one complete normalized phase period and a backward cycle by transporting $I_{\mathrm{e}}$ through one complete normalized phase period in the reverse direction, then compare the returned volumes with their corresponding endpoints~\cite{dwibedi2019temporal}. The resulting cycle loss is:
\begin{equation}
\mathcal{L}_{\mathrm{cyc}}
=
\mathcal{L}_{\mathrm{ncc}}(\hat{I}_{\mathrm{s}}^{\mathrm{cyc}}, I_{\mathrm{s}})
+
\mathcal{L}_{\mathrm{ncc}}(\hat{I}_{\mathrm{e}}^{\mathrm{cyc}}, I_{\mathrm{e}}),
\end{equation}
where $\hat{I}_{\mathrm{s}}^{\mathrm{cyc}}$ denotes the volume obtained by forward warping $I_{\mathrm{s}}$ through one full normalized phase period, and $\hat{I}_{\mathrm{e}}^{\mathrm{cyc}}$ denotes the volume obtained by backward warping $I_{\mathrm{e}}$ through one full normalized phase period.

After bidirectional transport, the two candidates are fused using distance-aware weighting on the observation-time axis and then refined by a lightweight residual network to compensate for appearance errors that are not fully resolved by deformation alone. The final prediction is written as:
\begin{equation}
\hat{I}^{t}
=
\frac{t_{\mathrm{e}}-t}{t_{\mathrm{e}}-t_{\mathrm{s}}}\hat{I}_{\mathrm{s}}^{\,t}
+
\frac{t-t_{\mathrm{s}}}{t_{\mathrm{e}}-t_{\mathrm{s}}}\hat{I}_{\mathrm{e}}^{\,t}
+
R\!\left(
[\hat{I}_{\mathrm{s}}^{\,t}, \hat{I}_{\mathrm{e}}^{\,t}]
\right),
\label{eq:residual_refinement_34}
\end{equation}
where $R(\cdot)$ takes the concatenation of the two warped candidates as input and predicts a local correction term.

Since refinement operates on the fused prediction rather than on each directional warp separately, the refinement loss is imposed only at the final output stage:
\begin{equation}
\mathcal{L}_{\mathrm{refine}}
=
\frac{1}{|\mathcal{T}|}
\sum_{t \in \mathcal{T}}
\left[
\mathcal{L}_{\mathrm{charb}}(\hat{I}^{t}, I_t)
+
\lambda_{\mathrm{grad}}
\mathcal{L}_{\mathrm{grad}}(\hat{I}^{t}, I_t)
\right],
\label{eq:refine_loss_34}
\end{equation}
where $\mathcal{L}_{\mathrm{grad}}$ denotes the gradient reconstruction loss, and $\lambda_{\mathrm{grad}}$ controls its contribution relative to the Charbonnier term.

Combining the deformation-level supervision, the refinement supervision, and the spectral regularization, the overall training objective is:
\begin{equation}
\mathcal{L}
=
\lambda_{\mathrm{morph}}\mathcal{L}_{\mathrm{morph}}
+
\lambda_{\mathrm{cyc}}\mathcal{L}_{\mathrm{cyc}}
+
\lambda_{\mathrm{refine}}\mathcal{L}_{\mathrm{refine}}
+
\lambda_{\mathrm{reg}}\mathcal{L}_{\mathrm{reg}}.
\label{eq:overall_loss_34}
\end{equation}

In this way, the deformation process is first constrained directly through bidirectional warping and phase-cycle consistency, after which residual refinement further improves appearance fidelity at the image level.

\begin{table*}[t]
\centering
\caption{Quantitative comparison on the ACDC and 4D-Lung datasets in terms of PSNR, NMI and SSIM, with model size (\#P) also reported.}
\label{tab:comparison}
\setlength{\tabcolsep}{3pt}
\renewcommand{\arraystretch}{1.1}

\newcommand{\numcolwcomp}{1.85cm}
\newcommand{\venuecolwcomp}{1.6cm}
\newcommand{\pcolwcomp}{1.1cm}
\newcolumntype{C}[1]{>{\centering\arraybackslash}m{#1}}

\begin{tabular}{l C{\venuecolwcomp} C{\numcolwcomp} C{\numcolwcomp} C{\numcolwcomp} C{\numcolwcomp} C{\numcolwcomp} C{\numcolwcomp} C{\pcolwcomp}}
\toprule
\multirow{2}{*}{Method} & \multirow{2}{*}{Venue}
& \multicolumn{3}{c}{ACDC}
& \multicolumn{3}{c}{4D-Lung}
& \multirow{2}{*}{\makecell[c]{\#P\\(M)}} \\
\cmidrule(lr){3-5} \cmidrule(lr){6-8}
& & PSNR $_{\text{dB}}^{\uparrow}$ & NMI $_{\times 10^{-2}}^{\uparrow}$ & SSIM $_{\times 10^{-2}}^{\uparrow}$
& PSNR $_{\text{dB}}^{\uparrow}$ & NMI $_{\times 10^{-2}}^{\uparrow}$ & SSIM $_{\times 10^{-2}}^{\uparrow}$ & \\
\midrule
VM~\cite{balakrishnan2019voxelmorph} & TMI'19
& 25.694 $_{\pm 0.074}$ & 53.543 $_{\pm 0.367}$ & 94.678 $_{\pm 0.104}$
& 26.077 $_{\pm 0.066}$ & 43.982 $_{\pm 0.355}$ & 86.055 $_{\pm 0.292}$ & 10.044 \\
TM~\cite{chen2022transmorph} & MedIA'22
& 27.125 $_{\pm 0.200}$ & 64.628 $_{\pm 0.671}$ & 96.644 $_{\pm 0.123}$
& 26.683 $_{\pm 0.051}$ & 48.945 $_{\pm 0.104}$ & 84.779 $_{\pm 0.125}$ & 10.804 \\
MPVF~\cite{wei2023mpvf} & JBHI'23
& 32.144 $_{\pm 0.011}$ & 70.658 $_{\pm 0.038}$ & 98.669 $_{\pm 0.003}$
& 29.909 $_{\pm 0.035}$ & 53.074 $_{\pm 0.389}$ & 90.180 $_{\pm 0.053}$ & 14.533 \\
SVIN~\cite{guo2020spatiotemporal} & CVPR'20
& 28.434 $_{\pm 0.443}$ & 60.178 $_{\pm 0.335}$ & 97.352 $_{\pm 0.152}$
& 27.312 $_{\pm 0.280}$ & 43.171 $_{\pm 0.392}$ & 85.442 $_{\pm 0.286}$ & 12.393 \\
IFRNet~\cite{kong2022ifrnet} & CVPR'22
& 32.447 $_{\pm 0.070}$ & 70.736 $_{\pm 0.149}$ & 98.741 $_{\pm 0.023}$
& 29.825 $_{\pm 0.058}$ & 51.567 $_{\pm 0.539}$ & 90.370 $_{\pm 0.092}$ & 24.520 \\
AMT~\cite{li2023amt} & CVPR'23
& 31.009 $_{\pm 0.267}$ & 71.461 $_{\pm 0.438}$ & 98.389 $_{\pm 0.077}$
& 28.881 $_{\pm 0.773}$ & 62.101 $_{\pm 1.080}$ & 87.954 $_{\pm 1.723}$ & 23.336\\
VFIFME~\cite{yu2023range} & CVPR'23
& 29.627 $_{\pm 3.297}$ & 66.215 $_{\pm 8.075}$ & 97.250 $_{\pm 4.624}$
& 29.711 $_{\pm 1.183}$ & 50.843 $_{\pm 4.208}$ & 89.677 $_{\pm 1.887}$ & 24.089 \\
UVI-Net~\cite{kim2024data} & CVPR'24
& 31.903 $_{\pm 0.118}$ & 72.252 $_{\pm 0.166}$ & 98.618 $_{\pm 0.027}$
& 28.948 $_{\pm 0.277}$ & 55.157 $_{\pm 0.841}$ & 88.750 $_{\pm 0.668}$ & 14.273 \\
PerVFI~\cite{wu2024perception} & CVPR'24
& 32.300 $_{\pm 0.702}$ & 68.755 $_{\pm 2.484}$ & 98.642 $_{\pm 0.599}$
& 28.523 $_{\pm 0.187}$ & 47.701 $_{\pm 0.332}$ & 87.614 $_{\pm 0.047}$ & 29.432 \\
FB-Diff~\cite{you2025fb} & ICCV'25
& 29.387 $_{\pm 0.103}$ & 62.135 $_{\pm 0.274}$ & 97.483 $_{\pm 0.163}$
& 27.617 $_{\pm 0.867}$ & 49.275 $_{\pm 1.345}$ & 80.913 $_{\pm 3.712}$ & 33.527 \\
DDM~\cite{kim2022diffusion} & MICCAI'22
& 28.558 $_{\pm 1.679}$ & 64.142 $_{\pm 1.349}$ & 97.237 $_{\pm 0.872}$
& 26.896 $_{\pm 0.101}$ & 42.544 $_{\pm 0.894}$ & 85.229 $_{\pm 0.270}$ & 23.178 \\
LDDM~\cite{chen2024ultrasound} & MICCAI'24
& 31.839 $_{\pm 0.033}$ & 72.333 $_{\pm 0.302}$ & 98.551 $_{\pm 0.013}$
& 27.563 $_{\pm 0.111}$ & 56.577 $_{\pm 8.077}$ & 83.499 $_{\pm 12.092}$ & 15.545 \\
TMSDF~\cite{zhang2025temporal} & MICCAI'25
& 31.974 $_{\pm 0.225}$ & 66.401 $_{\pm 0.833}$ & 98.558 $_{\pm 0.072}$
& 29.848 $_{\pm 0.023}$ & 48.360 $_{\pm 0.113}$ & 90.289 $_{\pm 1.086}$ & 42.283 \\
\midrule
\textbf{Ours} & -
& \textbf{32.503} $_{\pm 0.176}$ & \textbf{72.408} $_{\pm 0.780}$ & \textbf{98.765} $_{\pm 0.042}$
& \textbf{30.460} $_{\pm 0.038}$ & \textbf{62.659} $_{\pm 0.565}$ & \textbf{92.229} $_{\pm 0.141}$ & 20.813\\
\bottomrule
\end{tabular}
\end{table*}

\section{Experiments and Results}

\subsection{Experimental Settings}

We evaluate the proposed method on two public 4D medical imaging benchmarks, ACDC~\cite{bernard2018deep} and 4D-Lung~\cite{hugo2016_4dlung}. ACDC contains 150 cardiac MRI cases, split into 80, 20, and 50 cases for training, validation, and testing. All volumes are resized to $160 \times 160 \times 16$, and the end-diastolic and end-systolic phases are used as endpoints. The 4D-Lung dataset contains 500 cone-beam CT volumes, split at the patient level into 306, 84, and 110 cases to avoid leakage across subsets. All volumes are resized to $128 \times 128 \times 32$, and the $0\%$ and $50\%$ respiratory phases are used as endpoints. For both datasets, the endpoint times are set to $t_{\mathrm{s}}=0$ and $t_{\mathrm{e}}=0.5$. All volumes are histogram-equalized and linearly normalized to $[0,1]$.

We report Peak Signal-to-Noise Ratio (PSNR), Normalized Mutual Information (NMI), and Structural Similarity (SSIM) as the evaluation metrics. These metrics assess interpolation quality from complementary perspectives, including reconstruction fidelity, statistical dependence, and structural consistency. During both training and inference, the task is formulated in an endpoint-conditioned manner. Intermediate frames are used only as supervision and for quantitative evaluation, while the model input always consists of the two endpoint volumes together with the queried time.

For fair comparison, all methods are trained under the same optimization protocol. We use AdamW with an initial learning rate of $2 \times 10^{-4}$ and a cosine annealing schedule. Training is conducted for 500 epochs on ACDC and 200 epochs on 4D-Lung with early stopping. The batch size is set to 1, and gradient accumulation over 8 iterations is adopted to stabilize 3D training. We report the mean and standard deviation computed over multiple runs with different random seeds.

For the proposed method, the Fourier truncation order is set to $N=4$. We set $\lambda_{\mathrm{charb}}=10$, use an NCC window size of 7, set $\lambda_{\mathrm{grad}}=1.0$, and set the overall loss weights to $\lambda_{\mathrm{morph}}=1$, $\lambda_{\mathrm{cyc}}=1$, $\lambda_{\mathrm{refine}}=5$, and $\lambda_{\mathrm{reg}}=0.005$. For spectral regularization, the order exponent is set to $\gamma=1.5$.

\subsection{Comparison Study}

\begin{figure*}[tp]
\includegraphics[
width=0.9\textwidth,
height=4.5cm
]{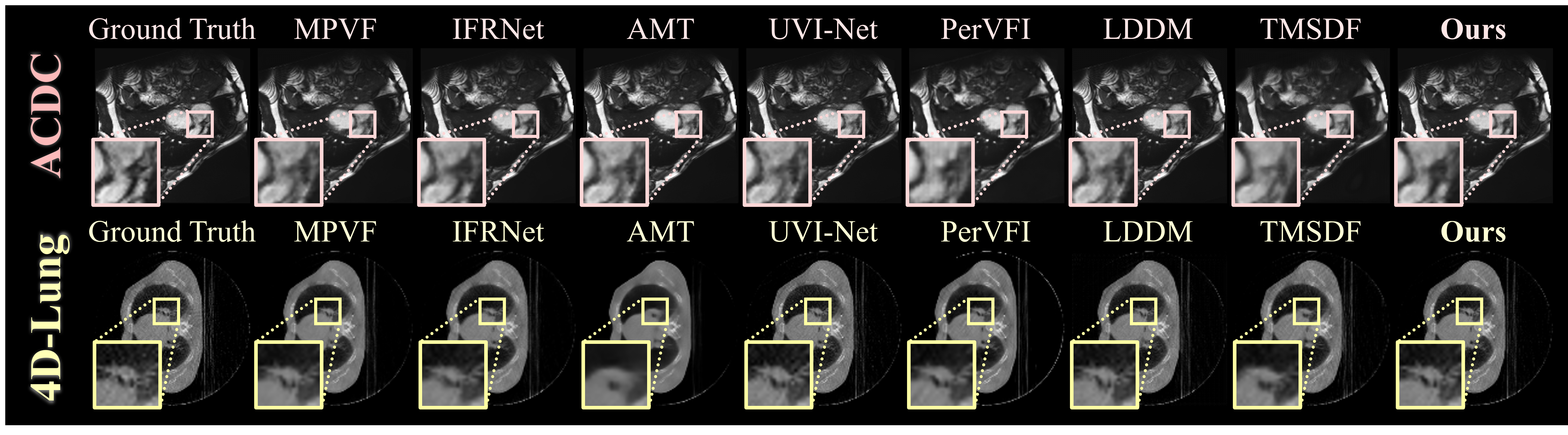}
\caption{Qualitative comparison on the ACDC and 4D-Lung datasets, showing better preservation of thin intracavitary low-signal structures in ACDC and clearer perihilar vessel-bundle-like structures in 4D-Lung.}
\label{fig:comparison}
\Description{Qualitative comparison of interpolated 4D medical images on two datasets. The top row shows ACDC and the bottom row shows 4D-Lung. Each row contains nine columns labeled, from left to right, Ground Truth, MPVF, IFRNet, AMT, UVI-Net, PerVFI, LDDM, TMSDF, and Ours, with “Ours” shown in bold. Large vertical dataset labels appear on the left side.

In the ACDC row, each panel shows a cardiac MRI slice. A light pink square marks a small region near the ventricular cavity, and a larger zoomed inset at the lower left is connected to the marked region by pink dotted guide lines. The enlarged comparison highlights a thin low-signal intracavitary structure. In the ground-truth inset, this structure appears narrow and continuous. Several competing methods make this local pattern appear blurrier, thicker, or less continuous, while the proposed method preserves a thinner and more coherent structure that is visually closer to the ground truth.

In the 4D-Lung row, each panel shows a thoracic CT slice. A light yellow square marks a region near the lung hilum, and a larger zoomed inset at the lower left is connected by yellow dotted guide lines. The enlarged region emphasizes fine perihilar vessel-bundle-like anatomy. In the ground-truth inset, the local texture appears relatively clear and structured. Several competing methods look smoother or less distinct in this region, whereas the proposed method shows a sharper and more organized local pattern that is visually closer to the ground truth.

Overall, the figure highlights local anatomical fidelity through matched zoomed insets, showing that the proposed method better preserves subtle intracavitary structures on ACDC and clearer perihilar vessel-bundle-like structures on 4D-Lung.}
\end{figure*}

Table~\ref{tab:comparison} shows clear performance differences among the compared methods on both ACDC and 4D-Lung. Earlier baselines such as VoxelMorph (VM)~\cite{balakrishnan2019voxelmorph} and TransMorph (TM)~\cite{chen2022transmorph} remain comparatively limited, suggesting that endpoint-conditioned interpolation requires stronger dynamic modeling than direct endpoint correspondence transfer. Dedicated volumetric interpolation methods, including SVIN~\cite{guo2020spatiotemporal}, MPVF~\cite{wei2023mpvf}, UVI-Net~\cite{kim2024data}, and TMSDF~\cite{zhang2025temporal}, achieve stronger results, while natural video interpolation models such as IFRNet~\cite{kong2022ifrnet} and AMT~\cite{li2023amt} remain competitive on several metrics. Generative variants including DDM~\cite{kim2022diffusion} and FB-Diff~\cite{you2025fb} also benefit from richer temporal priors, although their gains are less consistent across datasets and criteria. In contrast, our method achieves the most consistent overall performance on both benchmarks. This indicates that modeling near-periodic physiological motion through finite-Fourier deformation and temporal reparameterization improves 4D interpolation across datasets and metrics.

Fig.~\ref{fig:comparison} further shows visually meaningful improvements in interpolation quality. On ACDC, the predicted intermediate volumes better preserve regular boundaries and thin intracavitary low-signal structures, especially when local motion is less uniform. On 4D-Lung, the recovered intermediates better preserve coherent global organization and clearer perihilar vessel-bundle-like structures across time. In contrast, deformation-transfer and feature-based baselines more often show local inconsistency or unstable structural arrangement, while generative models may trade structural stability for visual smoothness. Together with Table~\ref{tab:comparison}, these observations indicate that the proposed method improves not only target-volume accuracy, but also the continuity and plausibility of the recovered motion.

\begin{figure*}[tp]
\includegraphics[
width=0.9\textwidth,
height=4cm
]{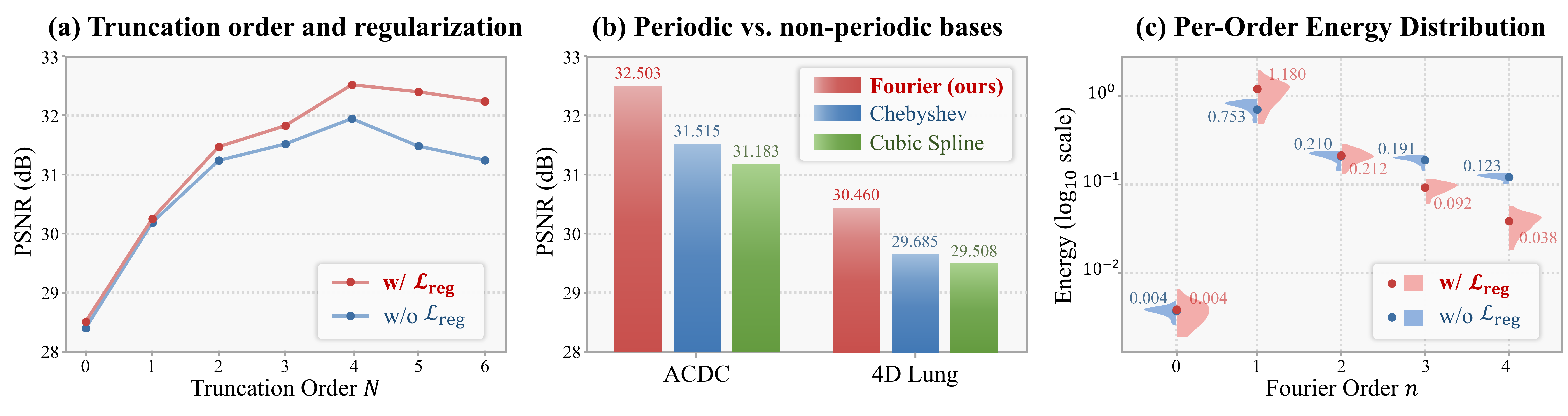}
\caption{Ablation studies on periodic deformation field modeling. 
(a) Effect of the truncation order $N$ and the regularization term $\mathcal{L}_{\mathrm{reg}}$. 
(b) Replacing the Fourier basis with Chebyshev polynomials and cubic splines to assess the role of periodic parameterization.
(c) Spectral energy distribution across Fourier orders, with and without $\mathcal{L}_{\mathrm{reg}}$.}
\label{fig:ablation}
\Description{A figure with three subplots arranged horizontally, labeled (a), (b), and (c), presenting ablation studies on periodic deformation field modeling.

Subplot (a), on the left, is a line chart showing the effect of the Fourier truncation order $N$ and the regularization term $\mathcal{L}_{\mathrm{reg}}$. The horizontal axis is the truncation order, increasing from left to right, and the vertical axis is interpolation performance. Two curves are shown: a red curve for the model with $\mathcal{L}_{\mathrm{reg}}$ and a blue curve for the model without $\mathcal{L}_{\mathrm{reg}}$. Both curves improve substantially as the order increases from low values to middle values, then level off or slightly decline at higher orders. Across most orders, the red curve stays above the blue curve, indicating better performance when the regularization term is used.

Subplot (b), in the middle, is a grouped bar chart comparing three periodic parameterizations: Fourier (ours), Chebyshev, and Cubic Spline. Two groups of bars are shown, corresponding to two datasets or evaluation settings. In both groups, the red Fourier bars are the tallest, the blue Chebyshev bars are lower, and the green Cubic Spline bars are slightly lower still. Numerical values are printed above the bars, showing that Fourier achieves the best results in both groups.

Subplot (c), on the right, shows the spectral energy distribution across Fourier orders, comparing the settings with and without $\mathcal{L}_{\mathrm{reg}}$. The horizontal axis is the Fourier order, and the vertical axis is the normalized energy magnitude. Red markers and shaded shapes indicate the model with $\mathcal{L}_{\mathrm{reg}}$, and blue markers and shaded shapes indicate the model without it. The displayed values show that lower orders contain the largest energy, while higher orders have progressively smaller energy. The two settings allocate energy differently across orders: with $\mathcal{L}_{\mathrm{reg}}$, the first order has noticeably higher energy, while some middle and higher orders show different relative magnitudes compared with the model without regularization.

Overall, the figure shows that using a Fourier parameterization gives the strongest results among the tested basis choices, that moderate truncation orders perform best, and that the regularization term changes the spectral distribution while also improving overall interpolation performance.}
\end{figure*}

\subsection{Ablation on Periodic Deformation Modeling}

\noindent\textbf{Effect of the Fourier truncation order.} Fig.~\ref{fig:ablation}(a) examines how the truncation order affects the proposed periodic deformation model. The overall trend shows that introducing a small number of phase-varying Fourier terms already brings clear improvement over the lowest-order setting, and that performance peaks at a moderate order before saturating or slightly declining. This suggests that the main interpolation dynamics can be represented within a compact low-order periodic subspace, whereas adding more orders mainly increases modeling flexibility in ways that are less tightly constrained by endpoint-conditioned supervision. The version with $\mathcal{L}_{\mathrm{reg}}$ remains consistently better across different orders, indicating that the proposed formulation benefits not only from periodic parameterization itself, but also from explicitly shaping how phase-dependent motion is distributed across spectral components.

\noindent\textbf{Comparison with alternative temporal basis functions.} Fig.~\ref{fig:ablation}(b) further compares the Fourier basis with two non-periodic alternatives, namely Chebyshev polynomials and cubic splines, on both ACDC and 4D-Lung. These alternatives also provide smooth continuous parameterization and achieve reasonably strong results, but the Fourier basis remains consistently better across the two datasets. This comparison suggests that the gain is not simply due to replacing discrete timestamps with another continuous basis. Rather, periodic parameterization offers a more suitable inductive bias by restricting the admissible deformation family to follow cycle-consistent evolution, which better matches the near-periodic nature of physiological motion.

\noindent\textbf{Distribution of spectral energy across Fourier orders.} To further inspect the learned representation, Fig.~\ref{fig:ablation}(c) reports the order-wise coefficient energy, computed as $E_n=|\Omega|^{-1}(\|a_n\|_2^2+\|b_n\|_2^2)$, with $E_0=|\Omega|^{-1}\|c\|_2^2$. With $\mathcal{L}_{\mathrm{reg}}$, energy is concentrated more heavily in the low orders, especially the first harmonic, while higher-order components are suppressed. Without regularization, the spectrum is flatter, with more energy remaining in the upper orders. The nearly unchanged zeroth-order term suggests that $\mathcal{L}_{\mathrm{reg}}$ reorganizes the phase-dependent spectrum rather than merely shrinking the overall deformation field. Together with Fig.~\ref{fig:ablation}(a), this supports capturing the dominant motion cycle with a compact periodic representation rather than dispersed higher-order variation.

\subsection{Ablation on Temporal Reparameterization}

\begin{figure}[tp]
\includegraphics[
width=0.4\textwidth,
height=7.5cm
]{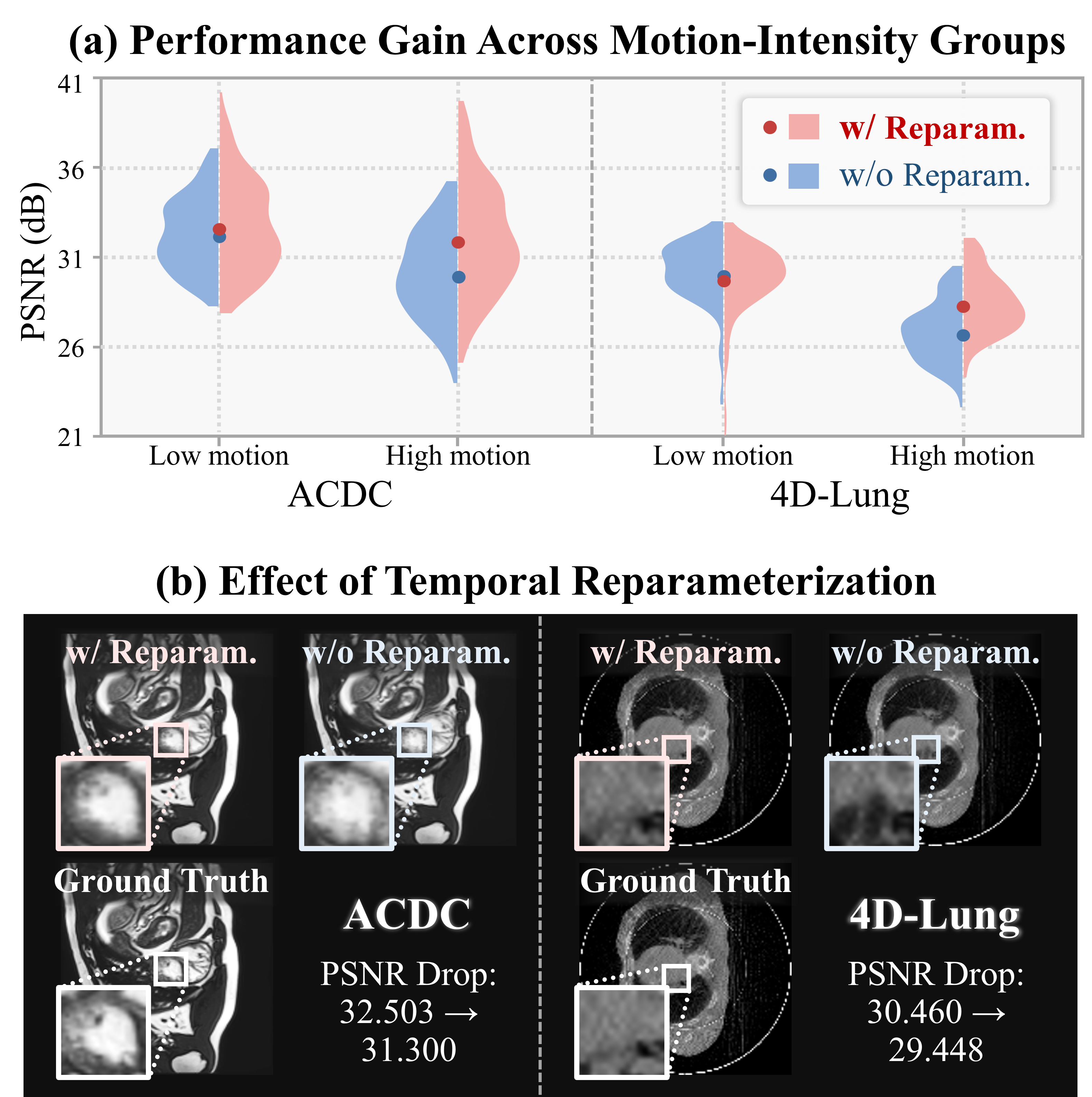}
\caption{Ablation on temporal reparameterization. 
(a) Analysis of interpolation performance across frame groups with different motion intensities on ACDC and 4D-Lung.
(b) Qualitative and quantitative comparison with and without temporal reparameterization, including visualization examples under large motion.}
\label{fig:ablation2}
\Description{Ablation figure on temporal reparameterization, consisting of two subfigures labeled (a) and (b).

Subfigure (a) shows four split violin plots comparing interpolation performance with and without temporal reparameterization under different motion intensities on ACDC and 4D-Lung. The first and second plots correspond to ACDC, and the third and fourth plots correspond to 4D-Lung. In each dataset, the first plot is for low-motion and the second plot is for high-motion. Red denotes results with temporal reparameterization, and blue denotes results without temporal reparameterization. Each violin includes a colored dot indicating the average performance. In the low-motion groups, the red and blue distributions are close, indicating that temporal reparameterization brings only a small difference. In the high-motion groups, the red distributions and mean markers are clearly higher than the blue ones, showing that temporal reparameterization provides a more obvious improvement when motion is larger.

Subfigure (b) presents qualitative and quantitative comparisons under large motion, with ACDC on the left and 4D-Lung on the right, separated by a vertical dashed divider. For each dataset, the image with temporal reparameterization and the image without temporal reparameterization are shown on the top row, and the ground-truth image is shown below. Small square boxes mark regions of interest, and enlarged insets connected by dotted guide lines highlight local details.

For ACDC, the result without temporal reparameterization is noticeably blurrier in the enlarged region, while the version with temporal reparameterization preserves clearer local structure and is visually closer to the ground truth. For 4D-Lung, the result without temporal reparameterization shows abnormal artifacts in the highlighted region, whereas the version with temporal reparameterization appears cleaner and more anatomically plausible, again closer to the ground truth.

Text annotations in subfigure (b) report the PSNR drop after removing temporal reparameterization: from 32.503 to 31.300 on ACDC, and from 30.460 to 29.448 on 4D-Lung. Overall, the figure shows that temporal reparameterization has limited effect in low-motion cases but yields clear quantitative and visual benefits in high-motion cases.}
\end{figure}

\noindent\textbf{Relation to frame-wise motion intensity.} Fig.~\ref{fig:ablation2}(a) analyzes how the effect of temporal reparameterization varies with motion intensity. We score each target frame by the mean $1-\mathrm{NCC}$ to its two neighboring frames, sort all frames by this score, and split them at the median into low-motion and high-motion groups. The resulting PSNR distributions show that the two variants remain close on the low-motion group, whereas a clearer gap appears on the high-motion group in both datasets. This is consistent with the role of temporal reparameterization: when inter-frame motion is limited, linear time already approximates motion progression reasonably well, while stronger motion makes such approximation less adequate and leaves more room for a phase-based parameterization to improve interpolation.

\noindent\textbf{Effect of temporal reparameterization.} Fig.~\ref{fig:ablation2}(b) compares the proposed model with and without temporal reparameterization on ACDC and 4D-Lung using visual examples and quantitative results. Consistent with the analysis above, the difference is clearer in large-motion cases, where removing temporal reparameterization leads to weaker interpolation quality and less accurate structural recovery. These results suggest that the gain comes from replacing linear time with a phase coordinate that better matches deformation progression.

\subsection{Additional Analysis}

\begin{table}[!t]
\centering
\caption{Loss term ablation on ACDC.}
\label{tab:ablation}

\newcommand{\termwclossabl}{0.65cm}
\newcommand{\numwclossabl}{1.15cm}

\begin{tabular}{%
>{\centering\arraybackslash}m{\termwclossabl}
>{\centering\arraybackslash}m{\termwclossabl}
>{\centering\arraybackslash}m{\termwclossabl}
>{\centering\arraybackslash}m{\termwclossabl}
|
>{\centering\arraybackslash}m{\numwclossabl}
>{\centering\arraybackslash}m{\numwclossabl}
>{\centering\arraybackslash}m{\numwclossabl}
}
\toprule
$\mathcal{L}_\mathrm{refine}$ & $\mathcal{L}_\mathrm{morph}$ & $\mathcal{L}_\mathrm{cycle}$ & $\mathcal{L}_\mathrm{reg}$ & PSNR$_{\text{dB}}^{\uparrow}$ & NMI$_{\times 10^{-2}}^{\uparrow}$ & SSIM$_{\times 10^{-2}}^{\uparrow}$ \\
\midrule
$\checkmark$ & & & & 30.700 & 63.753 & 98.206 \\
$\checkmark$ & $\checkmark$ & & & 31.786 & 66.766 & 98.528 \\
$\checkmark$ & $\checkmark$ & $\checkmark$ & & 31.927 & 67.812 & 98.583 \\
\midrule
$\checkmark$ & $\checkmark$ & $\checkmark$ & $\checkmark$ & 32.503 & 72.408 & 98.765 \\
\bottomrule
\end{tabular}
\end{table}

\noindent\textbf{Ablation on loss terms.} Table~\ref{tab:ablation} examines how different loss components contribute to the proposed framework. Starting from the basic reconstruction objective, each additional term brings further improvement across the reported metrics, and the full objective achieves the best overall results. This trend suggests that the gains do not come from a single constraint alone, but from the complementarity among the objectives. In particular, the cycle-related constraint helps improve temporal consistency of the bidirectional synthesis, while the regularization term further stabilizes the learned deformation representation. Rather than only improving the queried target volume itself, their combination also leads to a more coherent motion recovery process.

\begin{figure}[tp]
\includegraphics[
width=0.4\textwidth,
height=7.5cm
]{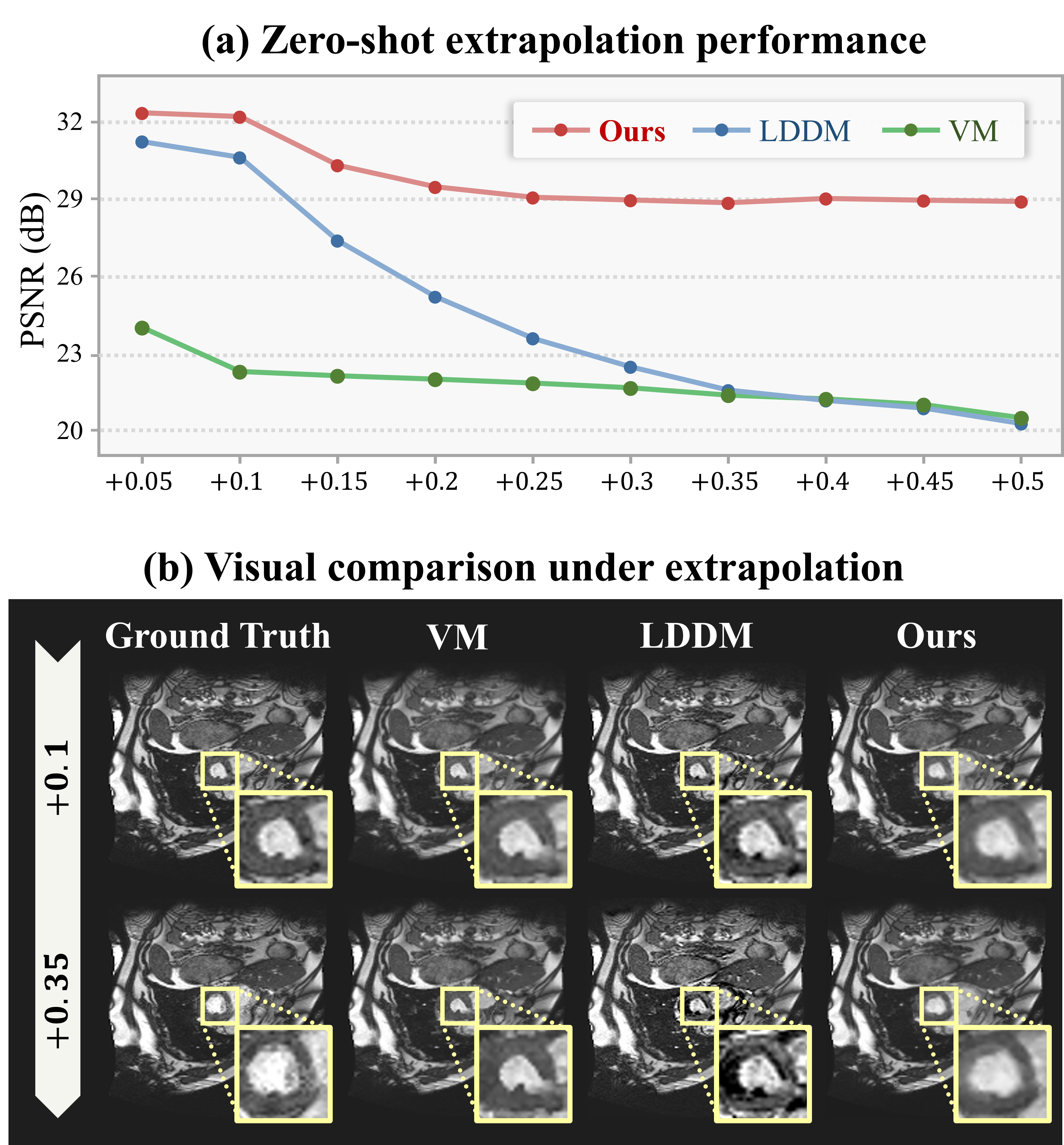}
\caption{Zero-shot extrapolation beyond the observed interval on the ACDC dataset. Models are trained using only frames within $[t_s, t_e]$ and directly evaluated on unseen future queries beyond the observed endpoint. (a) PSNR at different extrapolation offsets for our method and baselines. (b) Visual comparison at two increasing future query offsets.}
\label{fig:ablation3}
\Description{A figure with two vertically stacked parts illustrating unsupervised extrapolation beyond the observed interval on the ACDC dataset.

The upper part, labeled as subplot (a) by the caption, is a line chart of PSNR versus extrapolation offset. The horizontal axis represents increasingly larger future query offsets beyond the observed endpoint, and the vertical axis represents PSNR. Three methods are compared: Ours in red, LDDM in blue, and VM in green. The red curve remains the highest across all extrapolation offsets and decreases only slightly as the extrapolation horizon increases. The blue curve starts relatively high but drops steadily and substantially with increasing extrapolation distance. The green curve is consistently lower than the red curve and changes more gradually, staying close to the blue curve at later offsets. Overall, the plot shows that the proposed method maintains the strongest and most stable extrapolation performance over the full range of future query offsets.

The lower part, labeled as subplot (b) by the caption, is a qualitative comparison grid with four columns and two rows. The columns are Ground Truth, VM, LDDM, and Ours. The two rows correspond to two unseen future query offsets, marked on a vertical arrow at the left as $+0.1$ for the upper row and $+0.35$ for the lower row. Each image is a cardiac MRI slice. In every panel, a small square marks a local region of interest near the ventricular cavity, and a larger zoomed inset connected by dotted guide lines shows this region in detail.

At both extrapolation offsets, the Ground Truth images show a compact bright cardiac structure with a surrounding boundary that remains smooth and anatomically coherent. The proposed method produces zoomed regions that visually remain closest to the Ground Truth in both shape and local intensity pattern. VM appears blurrier and less well defined, especially in the magnified views. LDDM is closer to the Ground Truth at the smaller extrapolation offset but shows stronger distortion, artifact, and local structural degradation at the larger extrapolation offset $+0.35$. The visual comparison therefore matches the quantitative trend in the line plot, showing that the proposed method preserves local anatomy more reliably as extrapolation moves farther beyond the observed interval.}
\end{figure}

\noindent\textbf{Zero-shot extrapolation beyond the observed interval.} Fig.~\ref{fig:ablation3} evaluates queries beyond the observed interval $[t_s, t_e]$ defined by the endpoint frames. At short extrapolation distances, all methods produce reasonable predictions, but their differences increase farther beyond $t_e$. Quantitatively, our method shows slower performance decay, while VoxelMorph remains at a lower level and LDDM drops more sharply with increasing distance. The visual results are consistent with this trend: VoxelMorph tends to continue the observed motion pattern even when the target anatomy evolves differently, whereas LDDM becomes less stable and introduces implausible structures at larger offsets. In contrast, our method remains better aligned with the unseen future targets, suggesting that the learned deformation representation captures a more extensible motion pattern rather than only fitting the observed interval.

\section{Conclusion}

In this paper, we presented a phase-aware framework for 4D medical image interpolation that explicitly models structured and non-uniform physiological motion. Rather than treating interpolation as direct intensity prediction or relying on loosely constrained temporal conditioning, our method learns a continuous anatomical deformation process. A finite Fourier parameterization encodes periodic deformation patterns in a compact and interpretable manner, while phase-aligned temporal reparameterization bridges physical time and motion progression when the dynamics evolve unevenly. Combined with bidirectional endpoint warping and lightweight refinement, the framework enables anatomically plausible synthesis at arbitrary queried times. Experiments on ACDC and 4D-Lung demonstrate the effectiveness of the proposed design and suggest that explicitly modeling physiological motion in deformation space is a promising direction for continuous 4D medical image interpolation.

\begin{acks}
This work was supported in part by the National Natural Science
Foundation of China under Grant No.~62401246 and by the Shenzhen
Science and Technology Program under Grant Nos.~
JCYJ20250604185805008 and JCYJ20240813095112017.
The authors gratefully acknowledge Beijing Tiromu Medical Technology
Co., Ltd. and Shenzhen DE Sci\&Tech Co., Ltd. for their support in
technical validation and for providing a practical platform for this work.
\end{acks}

\bibliographystyle{ACM-Reference-Format}
\bibliography{refers}

\end{document}